\documentclass[11pt,a4paper]{article}

\usepackage[utf8]{inputenc}
\usepackage[T1]{fontenc}
\usepackage{lmodern}
\usepackage{amsmath,amssymb,amsfonts,bm}
\usepackage{geometry}
\usepackage{booktabs}
\usepackage{hyperref}
\usepackage{microtype}
\usepackage{mathtools}
\usepackage{enumitem}

\hypersetup{colorlinks=true, linkcolor=blue, citecolor=blue, urlcolor=blue}

\title{Learning Dynamics Reveal a Hierarchy of Weight-Induced Layerwise Gram Metrics}
\author{Claudio Nordio\\[0.8em]\large Draft research note\thanks{Circulated for discussion and feedback; comments are welcome at \texttt{c.nordio@gmail.com}.}}
\date{\today}

\newcommand{\odotp}{\odot}
\newcommand{\diag}{\operatorname{diag}}

\newcommand{\cU}{\mathcal U}
\newcommand{\cM}{\mathcal M}
\newcommand{\cF}{\mathcal F}
\newcommand{\cP}{\mathcal P}
\newcommand{\Id}{\mathrm{Id}}

\begin{document}

\maketitle

\begin{abstract}
We study feed-forward ReLU networks with fixed readout and quadratic loss, and rewrite gradient descent as a collective dynamics of activation fields and conjugate fields on the training set. Working to first order in the learning rate inside a fixed activation chamber, we derive explicitly the one-, two- and three-hidden-layer cases, and then give the arbitrary-depth recursion. For one hidden layer the activation dynamics closes directly and the residual update is governed by the product of an input Gram matrix and a co-activation/backpropagation Gram matrix. For two hidden layers a conjugate field is required, but no nontrivial pullback Gram metric has yet appeared. For three hidden layers the first weight-induced pullback Gram metric enters the conjugate-field dynamics. At arbitrary depth, activation variations propagate forward through a recursive response operator \(\cU_\ell^{\alpha\beta}\), while conjugate-field variations propagate backward through an effective transport operator \(\cM_\ell^{\alpha\beta}\). Their contractions reconstruct a layerwise residual kernel
\[
K_{\alpha\beta}^{(L)}=\sum_{\ell=1}^{L}Q_{\alpha\beta}^{(\ell-1)}S_{\alpha\beta}^{(\ell)}.
\]
The resulting description exposes a duality between push-forward and pullback transport across every layer cut, and identifies the first Gram metrics as the lowest nontrivial terms in a broader hierarchy of activation-conditioned transport operators. We deliberately stop at the level of collective fields, conjugate fields, residual kernels and cut-wise transport metrics, leaving the later tensorial geometric formulation outside the scope of this paper.
\end{abstract}

\section{Introduction}
\label{sec:introduction}

Gradient descent in neural networks is usually formulated as a dynamical system in weight space. The trainable variables are the weight matrices, while activations, predictions and residuals are derived quantities. This is the natural optimization viewpoint, but it is not always the most transparent way to describe the internal organization produced by learning.

This paper develops a complementary formulation for feed-forward ReLU networks with fixed readout. The central idea is to rewrite gradient descent as a dynamics of \emph{collective fields} defined on the training set: the activation fields \(u_\ell^\alpha\), their conjugate or backpropagated fields \(b_\ell^\alpha\), and the induced transport operators through which variations of these fields propagate across depth. The formulation is local in training time: throughout the derivation we work to first order in the learning rate \(\eta\), inside a fixed ReLU activation chamber. The masks are therefore held fixed during an infinitesimal update.

The motivation comes from the simplest case. With one hidden layer, the weight update can be eliminated from the activation dynamics. The residuals obey a closed kernel equation whose kernel factorizes into an input overlap and a co-activation/backpropagation overlap. This resembles the spirit of NTK descriptions \cite{jacot2018ntk}, but the emphasis here is different: rather than freezing the kernel in a limiting regime, we track how the finite-depth collective fields and their transport operators appear directly from gradient descent.

Increasing depth reveals a structured hierarchy. For two hidden layers one must introduce conjugate fields, since updating the second weight matrix changes the backpropagated field at the first hidden layer. For three hidden layers, a new object appears: a weight-induced pullback Gram metric of the form
\begin{equation}
\label{eq:intro-G}
G_\ell^{\alpha\beta}
=
\left(W^{(\ell+1)}\right)^T
D_{\ell+1}^{\alpha\beta}
W^{(\ell+1)},
\end{equation}
where \(D_{\ell+1}^{\alpha\beta}=A_{\ell+1}^\alpha A_{\ell+1}^\beta\) is a co-activation projector. Deeper networks generate further iterates of the same pullback transport mechanism. Thus the first Gram metric is not an isolated curiosity: it is the first nontrivial member of a hierarchy.

The goal of this paper is to make this hierarchy explicit while keeping the derivation as concrete as possible. We first develop the calculations for one, two and three hidden layers. These examples show exactly when activation closure holds, when conjugate fields are needed, and when the first nontrivial pullback Gram term enters. We then collect the general arbitrary-depth formulas, including the forward response recursion, the backward transport recursion, the residual kernel and the cut-wise push-forward/pullback decomposition.

The present work is deliberately limited in scope. It isolates the collective-field dynamics and its dual transport representation, without moving to the later tensorial geometric formulation. In this sense, the paper can be read as a finite-depth algebraic foundation for subsequent geometric developments.

\section{Setup and Fixed-Chamber ReLU Dynamics}
\label{sec:setup}

We consider a feed-forward network with \(L\) hidden layers and width \(N\). Training examples are indexed by \(\alpha=1,\ldots,M\). The input example is \(u_0^\alpha\), and hidden activations are
\begin{equation}
\label{eq:activations}
u_\ell^\alpha
=
\phi\!\left(W^{(\ell)}u_{\ell-1}^\alpha\right),
\qquad
\phi(x)=\max(x,0),
\qquad
\ell=1,\ldots,L.
\end{equation}
Inside a fixed ReLU chamber we introduce the binary mask
\begin{equation}
\label{eq:masks}
a_\ell^\alpha=\mathbf 1(u_\ell^\alpha>0),
\qquad
A_\ell^\alpha=\diag(a_\ell^\alpha),
\end{equation}
and the masked one-step forward map
\begin{equation}
\label{eq:F-one-step}
F_\ell^\alpha:=A_\ell^\alpha W^{(\ell)},
\qquad
u_\ell^\alpha=F_\ell^\alpha u_{\ell-1}^\alpha.
\end{equation}

The readout is fixed. We denote it by \(b_L\), or by \(b_L^\alpha\) when it is useful to keep sample labels. In the scalar unit-readout convention one may take \(b_L^\alpha=\mathbf 1\). Predictions, residuals and loss are
\begin{equation}
\label{eq:output-residual-loss}
\hat Y^\alpha
=
\frac1N (b_L^\alpha)^T u_L^\alpha,
\qquad
r^\alpha=
\hat Y^\alpha-Y^\alpha,
\qquad
\mathcal L=\frac1M\sum_{\alpha=1}^M (r^\alpha)^2.
\end{equation}

The conjugate fields are defined by backward transport:
\begin{equation}
\label{eq:conjugate-def}
b_{\ell-1}^\alpha
=
\left(W^{(\ell)}\right)^T A_\ell^\alpha b_\ell^\alpha
=
\left(F_\ell^\alpha\right)^T b_\ell^\alpha,
\qquad
\ell=1,\ldots,L.
\end{equation}
They are the feature-space fields that contract with activation variations at a given cut. Indeed, for any \(m\in\{1,\ldots,L\}\),
\begin{equation}
\label{eq:cut-duality}
\boxed{
\hat Y^\alpha
=
\frac1N(b_L^\alpha)^T u_L^\alpha
=
\frac1N(b_m^\alpha)^T u_m^\alpha .
}
\end{equation}
This simple identity is the basic duality behind the whole construction: activations propagate forward, conjugate fields propagate backward, but their contraction represents the same scalar output at every cut.

We also define the activation overlaps and conjugate co-activation overlaps
\begin{equation}
\label{eq:Q-S-def}
Q_{\alpha\beta}^{(\ell)}
=
\frac1N (u_\ell^\alpha)^T u_\ell^\beta,
\qquad
S_{\alpha\beta}^{(\ell)}
=
\frac1N
\left(a_\ell^\alpha\odot b_\ell^\alpha\right)^T
\left(a_\ell^\beta\odot b_\ell^\beta\right),
\end{equation}
for \(\ell=1,\ldots,L\), and \(Q^{(0)}\) for the input Gram matrix. In the fixed-readout unit convention, \(S^{(L)}_{\alpha\beta}=N^{-1}(a_L^\alpha)^T a_L^\beta\).

Finally, we define co-activation projectors
\begin{equation}
\label{eq:D-def}
D_\ell^{\alpha\beta}=A_\ell^\alpha A_\ell^\beta,
\qquad
\ell=1,\ldots,L.
\end{equation}

\subsection{Push-forward and pullback operators}

For a matrix \(X\) acting on layer \(\ell\), define the cross-sample push-forward operator from layer \(\ell\) to layer \(\ell+1\) by
\begin{equation}
\label{eq:pushforward-def}
\boxed{
\cF_\ell^{\alpha\beta}[X]
=
F_{\ell+1}^\alpha X\left(F_{\ell+1}^\beta\right)^T
=
A_{\ell+1}^\alpha W^{(\ell+1)}X
\left(W^{(\ell+1)}\right)^T A_{\ell+1}^\beta .
}
\end{equation}
The corresponding pullback operator is
\begin{equation}
\label{eq:pullback-def}
\boxed{
\cP_\ell^{\alpha\beta}[X]
=
\left(F_{\ell+1}^\alpha\right)^T X F_{\ell+1}^\beta
=
\left(W^{(\ell+1)}\right)^T A_{\ell+1}^\alpha X A_{\ell+1}^\beta W^{(\ell+1)} .
}
\end{equation}
The first pullback Gram metric associated with layer \(\ell+1\) is
\begin{equation}
\label{eq:G-def}
G_\ell^{\alpha\beta}
=
\cP_\ell^{\alpha\beta}[\Id]
=
\left(W^{(\ell+1)}\right)^T
D_{\ell+1}^{\alpha\beta}
W^{(\ell+1)}.
\end{equation}
This object is positive semidefinite when \(\alpha=\beta\) and, more generally, acts as a cross-sample pullback of the co-activation geometry of layer \(\ell+1\) to layer \(\ell\).

\section{Gradient Descent in Collective Fields}
\label{sec:gd-collective}

The gradient of the loss with respect to \(W^{(\ell)}\) is obtained by differentiating \(\hat Y^\beta\) through the cut at layer \(\ell\). This gives the first-order update
\begin{equation}
\label{eq:weight-update-general}
\boxed{
\Delta W^{(\ell)}
=
-
\frac{2\eta}{NM}
\sum_{\beta=1}^{M}
\left(a_\ell^\beta\odotp b_\ell^\beta\right)
\left(u_{\ell-1}^\beta\right)^T
r^\beta .
}
\end{equation}
All formulas below follow from inserting this update into the variations of activations and conjugate fields while keeping the masks fixed.

\section{One Hidden Layer: Exact Activation Closure}
\label{sec:one-layer}

For one hidden layer,
\begin{equation}
 u_1^\alpha=A_1^\alpha W^{(1)}u_0^\alpha,
 \qquad
 \hat Y^\alpha=\frac1N(b_1^\alpha)^T u_1^\alpha,
\end{equation}
where \(b_1\) is the fixed readout. The weight update is
\begin{equation}
\label{eq:one-layer-weight}
\Delta W^{(1)}
=
-
\frac{2\eta}{NM}
\sum_\beta
\left(a_1^\beta\odotp b_1^\beta\right)
\left(u_0^\beta\right)^T
r^\beta .
\end{equation}
Therefore
\begin{align}
\Delta u_1^\alpha
&=
A_1^\alpha\Delta W^{(1)}u_0^\alpha
\nonumber\\
&=
-
\frac{2\eta}{NM}
\sum_\beta
A_1^\alpha
\left(a_1^\beta\odotp b_1^\beta\right)
\left(u_0^\beta\right)^T u_0^\alpha
r^\beta
\nonumber\\
&=
-
\frac{2\eta}{M}
\sum_\beta
Q_{\alpha\beta}^{(0)}D_1^{\alpha\beta}b_1^\beta r^\beta .
\end{align}
Thus the one-layer activation response is closed:
\begin{equation}
\label{eq:U1}
\boxed{
\Delta u_1^\alpha
=
-\frac{2\eta}{M}
\sum_\beta
\cU_1^{\alpha\beta}b_1^\beta r^\beta,
\qquad
\cU_1^{\alpha\beta}=Q_{\alpha\beta}^{(0)}D_1^{\alpha\beta}.
}
\end{equation}
Projecting on the readout gives
\begin{align}
\Delta r^\alpha
&=
\frac1N(b_1^\alpha)^T\Delta u_1^\alpha
\nonumber\\
&=
-\frac{2\eta}{M}
\sum_\beta
Q_{\alpha\beta}^{(0)}
\frac1N
\left(a_1^\alpha\odotp b_1^\alpha\right)^T
\left(a_1^\beta\odotp b_1^\beta\right)
r^\beta
\nonumber\\
&=
-\frac{2\eta}{M}
\sum_\beta
K_{\alpha\beta}^{(1)}r^\beta,
\end{align}
with
\begin{equation}
\label{eq:K-one-layer}
\boxed{
K_{\alpha\beta}^{(1)}=Q_{\alpha\beta}^{(0)}S_{\alpha\beta}^{(1)}.
}
\end{equation}
This is the simplest collective closure: no additional conjugate-field dynamics is needed beyond the fixed readout, and the residual kernel factorizes into an input Gram factor and a co-activation/backpropagation Gram factor.

\section{Two Hidden Layers: Emergence of the Conjugate Field}
\label{sec:two-layers}

For two hidden layers,
\begin{equation}
 u_1^\alpha=F_1^\alpha u_0^\alpha,
 \qquad
 u_2^\alpha=F_2^\alpha u_1^\alpha,
 \qquad
 b_1^\alpha=(F_2^\alpha)^T b_2^\alpha,
\end{equation}
where \(b_2\) is fixed. The first activation still satisfies
\begin{equation}
\label{eq:two-U1}
\Delta u_1^\alpha
=
-\frac{2\eta}{M}
\sum_\beta
Q_{\alpha\beta}^{(0)}D_1^{\alpha\beta}b_1^\beta r^\beta.
\end{equation}
The second activation receives two contributions: one from the direct update of \(W^{(2)}\), and one from the propagation of \(\Delta u_1\) through the current forward map:
\begin{equation}
\Delta u_2^\alpha
=
A_2^\alpha\Delta W^{(2)}u_1^\alpha
+
F_2^\alpha\Delta u_1^\alpha .
\end{equation}
The direct term is
\begin{align}
A_2^\alpha\Delta W^{(2)}u_1^\alpha
&=
-\frac{2\eta}{M}
\sum_\beta
Q_{\alpha\beta}^{(1)}D_2^{\alpha\beta}b_2^\beta r^\beta.
\end{align}
The propagated term is
\begin{align}
F_2^\alpha\Delta u_1^\alpha
&=
-\frac{2\eta}{M}
\sum_\beta
F_2^\alpha
\left(Q_{\alpha\beta}^{(0)}D_1^{\alpha\beta}\right)
b_1^\beta r^\beta
\nonumber\\
&=
-\frac{2\eta}{M}
\sum_\beta
Q_{\alpha\beta}^{(0)}
F_2^\alpha D_1^{\alpha\beta}(F_2^\beta)^T
b_2^\beta r^\beta
\nonumber\\
&=
-\frac{2\eta}{M}
\sum_\beta
\cF_1^{\alpha\beta}
\left[Q_{\alpha\beta}^{(0)}D_1^{\alpha\beta}\right]
b_2^\beta r^\beta.
\end{align}
Therefore
\begin{equation}
\label{eq:U2-explicit}
\boxed{
\Delta u_2^\alpha
=
-\frac{2\eta}{M}
\sum_\beta
\cU_2^{\alpha\beta}b_2^\beta r^\beta,
\qquad
\cU_2^{\alpha\beta}
=
Q_{\alpha\beta}^{(1)}D_2^{\alpha\beta}
+
\cF_1^{\alpha\beta}
\left[Q_{\alpha\beta}^{(0)}D_1^{\alpha\beta}\right].
}
\end{equation}

The new feature of two hidden layers is that the first conjugate field is dynamical. Since \(b_1^\alpha=(F_2^\alpha)^Tb_2^\alpha\),
\begin{align}
\Delta b_1^\alpha
&=
\left(\Delta W^{(2)}\right)^T A_2^\alpha b_2^\alpha
\nonumber\\
&=
-\frac{2\eta}{NM}
\sum_\beta
u_1^\beta
\left(a_2^\beta\odotp b_2^\beta\right)^T
A_2^\alpha b_2^\alpha
r^\beta
\nonumber\\
&=
-\frac{2\eta}{M}
\sum_\beta
S_{\alpha\beta}^{(2)}u_1^\beta r^\beta.
\end{align}
Thus
\begin{equation}
\label{eq:M1-two-layer}
\boxed{
\Delta b_1^\alpha
=
-\frac{2\eta}{M}
\sum_\beta
\cM_1^{\alpha\beta}u_1^\beta r^\beta,
\qquad
\cM_1^{\alpha\beta}=S_{\alpha\beta}^{(2)}\Id.
}
\end{equation}
No nontrivial pullback Gram metric has yet appeared: the effective backward operator is still a scalar multiple of the identity. The residual kernel is
\begin{equation}
\label{eq:K-two-layer}
\boxed{
K_{\alpha\beta}^{(2)}
=
Q_{\alpha\beta}^{(0)}S_{\alpha\beta}^{(1)}
+
Q_{\alpha\beta}^{(1)}S_{\alpha\beta}^{(2)}.
}
\end{equation}
At the cut \(m=1\), the first term is reconstructed by \((b_1)^T\Delta u_1\), while the second is reconstructed by \((u_1)^T\Delta b_1\). This is the first instance of the lower/upper cut decomposition developed below.

\section{Three Hidden Layers: First Pullback Gram Metric}
\label{sec:three-layers}

For three hidden layers,
\begin{equation}
 u_3^\alpha=F_3^\alpha F_2^\alpha F_1^\alpha u_0^\alpha,
 \qquad
 b_2^\alpha=(F_3^\alpha)^Tb_3^\alpha,
 \qquad
 b_1^\alpha=(F_2^\alpha)^Tb_2^\alpha.
\end{equation}
The forward activation responses are generated recursively. The first two are the same as before, and the third is
\begin{equation}
\label{eq:U3-explicit}
\boxed{
\begin{aligned}
\cU_3^{\alpha\beta}
&=
Q_{\alpha\beta}^{(2)}D_3^{\alpha\beta}
+
\cF_2^{\alpha\beta}
\left[Q_{\alpha\beta}^{(1)}D_2^{\alpha\beta}\right]
+
\cF_2^{\alpha\beta}
\left[
\cF_1^{\alpha\beta}
\left[Q_{\alpha\beta}^{(0)}D_1^{\alpha\beta}\right]
\right].
\end{aligned}
}
\end{equation}
Equivalently,
\begin{equation}
\Delta u_3^\alpha
=
-\frac{2\eta}{M}
\sum_\beta
\cU_3^{\alpha\beta}b_3^\beta r^\beta.
\end{equation}
Thus the activation response is a sum of direct update insertions at each layer, pushed forward to the output layer.

The conjugate-field dynamics now exposes the first nontrivial pullback Gram metric. At the second hidden layer,
\begin{equation}
\label{eq:M2-three}
\boxed{
\Delta b_2^\alpha
=
-\frac{2\eta}{M}
\sum_\beta
S_{\alpha\beta}^{(3)}u_2^\beta r^\beta,
\qquad
\cM_2^{\alpha\beta}=S_{\alpha\beta}^{(3)}\Id.
}
\end{equation}
At the first hidden layer, \(b_1^\alpha=(F_2^\alpha)^Tb_2^\alpha\), so \(\Delta b_1\) has a direct term from \(\Delta W^{(2)}\) and a propagated term from \(\Delta b_2\):
\begin{equation}
\Delta b_1^\alpha
=
(\Delta W^{(2)})^T A_2^\alpha b_2^\alpha
+
(F_2^\alpha)^T\Delta b_2^\alpha.
\end{equation}
The first term gives \(S_{\alpha\beta}^{(2)}\Id\), as in the two-layer case. For the second term,
\begin{align}
(F_2^\alpha)^T\Delta b_2^\alpha
&=
-\frac{2\eta}{M}
\sum_\beta
S_{\alpha\beta}^{(3)}
(F_2^\alpha)^T u_2^\beta r^\beta
\nonumber\\
&=
-\frac{2\eta}{M}
\sum_\beta
S_{\alpha\beta}^{(3)}
(F_2^\alpha)^T F_2^\beta u_1^\beta r^\beta
\nonumber\\
&=
-\frac{2\eta}{M}
\sum_\beta
S_{\alpha\beta}^{(3)}
G_1^{\alpha\beta}u_1^\beta r^\beta,
\end{align}
where
\begin{equation}
\label{eq:G1-three}
G_1^{\alpha\beta}
=
(F_2^\alpha)^T F_2^\beta
=
\left(W^{(2)}\right)^T
D_2^{\alpha\beta}
W^{(2)}.
\end{equation}
Hence
\begin{equation}
\label{eq:M1-three}
\boxed{
\Delta b_1^\alpha
=
-\frac{2\eta}{M}
\sum_\beta
\cM_1^{\alpha\beta}u_1^\beta r^\beta,
\qquad
\cM_1^{\alpha\beta}
=
S_{\alpha\beta}^{(2)}\Id
+
S_{\alpha\beta}^{(3)}G_1^{\alpha\beta}.
}
\end{equation}
This is the first depth at which a weight-induced pullback Gram metric is unavoidable in the conjugate-field dynamics. The residual kernel is
\begin{equation}
\label{eq:K-three-layer}
\boxed{
K_{\alpha\beta}^{(3)}
=
Q_{\alpha\beta}^{(0)}S_{\alpha\beta}^{(1)}
+
Q_{\alpha\beta}^{(1)}S_{\alpha\beta}^{(2)}
+
Q_{\alpha\beta}^{(2)}S_{\alpha\beta}^{(3)}.
}
\end{equation}
The three-layer case therefore separates the roles of the two hierarchies: the residual kernel still has a simple layerwise scalar form, while the internal conjugate-field dynamics already contains a nontrivial operator-valued pullback term.

\section{Arbitrary Depth: Recursive Collective-Field Closure}
\label{sec:arbitrary-depth}

The explicit low-depth calculations suggest the general pattern. At arbitrary depth, activation variations propagate forward through operators \(\cU_\ell^{\alpha\beta}\), while conjugate-field variations propagate backward through operators \(\cM_\ell^{\alpha\beta}\).

\subsection{Activation-field recursion}

For every hidden layer \(\ell=1,\ldots,L\),
\begin{equation}
\label{eq:activation-general}
\boxed{
\Delta u_\ell^\alpha
=
-\frac{2\eta}{M}
\sum_{\beta=1}^{M}
\cU_\ell^{\alpha\beta} b_\ell^\beta r^\beta .
}
\end{equation}
The forward response operators satisfy
\begin{equation}
\label{eq:U-recursion}
\boxed{
\cU_\ell^{\alpha\beta}
=
Q_{\alpha\beta}^{(\ell-1)}D_\ell^{\alpha\beta}
+
\cF_{\ell-1}^{\alpha\beta}
\left[\cU_{\ell-1}^{\alpha\beta}\right],
\qquad
\ell=2,\ldots,L,
}
\end{equation}
with initial condition
\begin{equation}
\label{eq:U-initial}
\boxed{
\cU_1^{\alpha\beta}=Q_{\alpha\beta}^{(0)}D_1^{\alpha\beta}.
}
\end{equation}
Thus \(\cU_\ell\) is obtained by inserting the direct layer-\(j\) update at each \(j\le \ell\) and pushing it forward to layer \(\ell\). Its first terms are
\begin{equation}
Q_{\alpha\beta}^{(\ell-1)}D_\ell^{\alpha\beta},
\qquad
Q_{\alpha\beta}^{(\ell-2)}\cF_{\ell-1}^{\alpha\beta}[D_{\ell-1}^{\alpha\beta}],
\qquad
Q_{\alpha\beta}^{(\ell-3)}\cF_{\ell-1}^{\alpha\beta}
\left[
\cF_{\ell-2}^{\alpha\beta}[D_{\ell-2}^{\alpha\beta}]
\right],
\end{equation}
with the evident continuation.

\subsection{Conjugate-field recursion}

For \(\ell=1,\ldots,L-1\), the conjugate-field variation can be written as
\begin{equation}
\label{eq:conjugate-general}
\boxed{
\Delta b_\ell^\alpha
=
-\frac{2\eta}{M}
\sum_{\beta=1}^{M}
\cM_\ell^{\alpha\beta}u_\ell^\beta r^\beta .
}
\end{equation}
The effective backward transport operators obey
\begin{equation}
\label{eq:M-recursion}
\boxed{
\cM_\ell^{\alpha\beta}
=
S_{\alpha\beta}^{(\ell+1)}\Id
+
\cP_\ell^{\alpha\beta}
\left[
\cM_{\ell+1}^{\alpha\beta}
\right],
\qquad
\ell=1,\ldots,L-2,
}
\end{equation}
with terminal condition
\begin{equation}
\label{eq:M-terminal}
\boxed{
\cM_{L-1}^{\alpha\beta}=S_{\alpha\beta}^{(L)}\Id.
}
\end{equation}
Iterating gives the hierarchy
\begin{equation}
\label{eq:M-hierarchy}
S^{(\ell+1)}\Id,
\qquad
S^{(\ell+2)}G_\ell,
\qquad
S^{(\ell+3)}\cP_\ell[G_{\ell+1}],
\qquad
S^{(\ell+4)}\cP_\ell[\cP_{\ell+1}[G_{\ell+2}]],
\qquad
\ldots
\end{equation}
The first nontrivial term is the pullback Gram metric \(G_\ell=\cP_\ell[\Id]\). Higher terms are generally not simple Gram matrices; they are iterated pullback transport operators.

\subsection{Residual-field dynamics}

Projecting the last activation variation on the fixed readout gives
\begin{equation}
\label{eq:residual-general}
\Delta r^\alpha
=
\frac1N(b_L^\alpha)^T\Delta u_L^\alpha
=
-\frac{2\eta}{M}
\sum_{\beta=1}^M
K_{\alpha\beta}^{(L)}r^\beta,
\end{equation}
where
\begin{equation}
\label{eq:K-general}
\boxed{
K_{\alpha\beta}^{(L)}
=
\sum_{\ell=1}^{L}
Q_{\alpha\beta}^{(\ell-1)}S_{\alpha\beta}^{(\ell)}.
}
\end{equation}
This is the residual kernel seen by the training errors. It remains a layerwise sum of products of activation overlaps and conjugate co-activation overlaps even though the internal field dynamics contains nontrivial operator-valued transports.

The kernel in \eqref{eq:K-general} should be distinguished from the internal operators \(\cU_\ell\) and \(\cM_\ell\). The scalar residual kernel sees only their contractions. The full field dynamics retains the transport structure that is invisible after projection onto the residuals.

\section{Cut Decomposition and Dual Transport Metrics}
\label{sec:cut-decomposition}

The output duality \eqref{eq:cut-duality} implies that at every cut \(m\),
\begin{equation}
\label{eq:cut-residual-delta}
\boxed{
\Delta r^\alpha
=
\frac1N(b_m^\alpha)^T\Delta u_m^\alpha
+
\frac1N(u_m^\alpha)^T\Delta b_m^\alpha .
}
\end{equation}
The first term reconstructs the contribution of layers below or at the cut, while the second reconstructs the contribution of layers above the cut:
\begin{align}
\label{eq:cut-lower-upper-1}
\frac1N(b_m^\alpha)^T\Delta u_m^\alpha
&=
-\frac{2\eta}{M}
\sum_\beta
\left[
\sum_{j=1}^{m}
Q_{\alpha\beta}^{(j-1)}S_{\alpha\beta}^{(j)}
\right]r^\beta,
\\
\label{eq:cut-lower-upper-2}
\frac1N(u_m^\alpha)^T\Delta b_m^\alpha
&=
-\frac{2\eta}{M}
\sum_\beta
\left[
\sum_{j=m+1}^{L}
Q_{\alpha\beta}^{(j-1)}S_{\alpha\beta}^{(j)}
\right]r^\beta.
\end{align}
Equivalently,
\begin{align}
\label{eq:U-contraction}
\frac1N(b_m^\alpha)^T\cU_m^{\alpha\beta}b_m^\beta
&=
\sum_{j=1}^{m}
Q_{\alpha\beta}^{(j-1)}S_{\alpha\beta}^{(j)},
\\
\label{eq:M-contraction}
\frac1N(u_m^\alpha)^T\cM_m^{\alpha\beta}u_m^\beta
&=
\sum_{j=m+1}^{L}
Q_{\alpha\beta}^{(j-1)}S_{\alpha\beta}^{(j)}.
\end{align}
Thus \(\cU_m\) constructs the lower part of the residual kernel, while \(\cM_m\) constructs the upper part. This is the precise algebraic meaning of the push-forward/pullback duality at a cut.

\subsection{Transport metrics representing upper and lower Grams at a cut}

For \(j\ge m\), define the sample-dependent forward transport from the cut \(m\) to layer \(j\) by
\begin{equation}
\label{eq:F-m-j}
F_{m\to j}^\alpha
=
F_j^\alpha F_{j-1}^\alpha\cdots F_{m+1}^\alpha,
\qquad
F_{m\to m}^\alpha=\Id.
\end{equation}
Then \(u_j^\alpha=F_{m\to j}^\alpha u_m^\alpha\). The cross-sample pullback metric at the cut is
\begin{equation}
\label{eq:T-cut}
\boxed{
T_{m\to j}^{\alpha\beta}
=
\left(F_{m\to j}^\alpha\right)^T F_{m\to j}^\beta .
}
\end{equation}
It represents the forward activation Gram at layer \(j\) by contraction at the cut:
\begin{equation}
\label{eq:T-contract}
\frac1N(u_m^\alpha)^T
T_{m\to j}^{\alpha\beta}
u_m^\beta
=
\frac1N(u_j^\alpha)^T u_j^\beta
=
Q_{\alpha\beta}^{(j)}.
\end{equation}
For the backward side, take \(j\le m\) and define
\begin{equation}
\label{eq:F-j-m}
F_{j\to m}^\alpha
=
F_m^\alpha F_{m-1}^\alpha\cdots F_{j+1}^\alpha,
\qquad
F_{m\to m}^\alpha=\Id.
\end{equation}
The push-forward representation of the co-activation/backpropagation Gram at layer \(j\) is
\begin{equation}
\label{eq:Sigma-cut}
\boxed{
\Sigma_{j\to m}^{\alpha\beta}
=
F_{j\to m}^\alpha
D_j^{\alpha\beta}
\left(F_{j\to m}^\beta\right)^T .
}
\end{equation}
Since \(b_j^\alpha=(F_{j\to m}^\alpha)^Tb_m^\alpha\), one has
\begin{equation}
\label{eq:Sigma-contract}
\frac1N(b_m^\alpha)^T
\Sigma_{j\to m}^{\alpha\beta}
b_m^\beta
=
\frac1N
(A_j^\alpha b_j^\alpha)^T(A_j^\beta b_j^\beta)
=
S_{\alpha\beta}^{(j)}.
\end{equation}
Consequently the two response operators have the explicit cut representations
\begin{equation}
\label{eq:M-cut-sum}
\boxed{
\cM_m^{\alpha\beta}
=
\sum_{j=m}^{L-1}
S_{\alpha\beta}^{(j+1)}
T_{m\to j}^{\alpha\beta},
}
\end{equation}
and
\begin{equation}
\label{eq:U-cut-sum}
\boxed{
\cU_m^{\alpha\beta}
=
\sum_{j=1}^{m}
Q_{\alpha\beta}^{(j-1)}
\Sigma_{j\to m}^{\alpha\beta}.
}
\end{equation}
Indeed, contracting \eqref{eq:M-cut-sum} with \(u_m^\alpha,u_m^\beta\) gives the upper kernel sum in \eqref{eq:M-contraction}, and contracting \eqref{eq:U-cut-sum} with \(b_m^\alpha,b_m^\beta\) gives the lower kernel sum in \eqref{eq:U-contraction}.

\section{Structural Consequences}
\label{sec:structural}

The derivation yields several structural consequences.

\paragraph{Layerwise residual kernel.}
For any depth, the residual dynamics closes on the residual vector through the kernel \eqref{eq:K-general}. Although internal fields require operator-valued transports, the projected training-error dynamics depends only on the scalar layerwise products \(Q^{(\ell-1)}S^{(\ell)}\).

\paragraph{First nontrivial Gram term.}
A nontrivial weight-induced pullback Gram metric first appears at three hidden layers. In the two-hidden-layer network the backward operator at the first hidden layer is \(S^{(2)}\Id\). In the three-hidden-layer network it becomes \(S^{(2)}\Id+S^{(3)}G_1\). This marks the first point at which deeper weights enter the conjugate-field dynamics as a genuine operator on an earlier feature space.

\paragraph{Push-forward/pullback duality.}
The operators \(\cU_m\) and \(\cM_m\) are dual. The former pushes lower co-activation seeds upward to the cut; the latter pulls upper activation Grams downward to the cut. Their contractions reconstruct complementary pieces of the same residual kernel.

\paragraph{Observable state variables.}
At the collective-field level, the natural variables are
\begin{equation}
\label{eq:obs-state}
\mathcal S_{\rm obs}
=
\left\{
 u_\ell^\alpha,
 b_\ell^\alpha,
 Q_{\alpha\beta}^{(\ell)},
 S_{\alpha\beta}^{(\ell)},
 \cU_\ell^{\alpha\beta},
 \cM_\ell^{\alpha\beta},
 \cF_\ell^{\alpha\beta},
 \cP_\ell^{\alpha\beta},
 G_\ell^{\alpha\beta}
\right\}.
\end{equation}
This is not a strict microscopic Markovian closure in the fields alone, because the vector updates still use the current forward maps. It is, however, a closed algebraic description of how gradient descent generates the residual kernel and the finite hierarchy of transport operators visible at each cut.

\section{Numerical Checks and Spectral Diagnostics}
\label{sec:numerics}

The identities above suggest direct numerical checks. During training, one can record \(u_\ell^\alpha\), \(b_\ell^\alpha\), \(Q^{(\ell)}\), \(S^{(\ell)}\), and verify that the observed residual update is reproduced by
\begin{equation}
\Delta r^\alpha
\simeq
-\frac{2\eta}{M}
\sum_\beta
\left(
\sum_{\ell=1}^{L}Q_{\alpha\beta}^{(\ell-1)}S_{\alpha\beta}^{(\ell)}
\right)r^\beta
\end{equation}
up to higher-order terms in \(\eta\) and mask changes. One can also test the cut identities \eqref{eq:cut-lower-upper-1}--\eqref{eq:cut-lower-upper-2} by computing the two sides independently.

A second diagnostic concerns the onset of the first pullback Gram term. Comparing two- and three-hidden-layer networks should reveal that \(\cM_1\) is scalar in the two-layer case but contains the operator \(G_1=(W^{(2)})^TD_2W^{(2)}\) in the three-layer case. More generally, the spectra of \(G_\ell^{\alpha\beta}\) and of the iterated pullback transports in \(\cM_\ell\) measure how deeper co-activation geometry is represented on earlier feature spaces.

This perspective is compatible with empirical spectral approaches such as WeightWatcher and Heavy-Tailed Self-Regularization \cite{martin2021heavytailed,martin_weightwatcher}. Those approaches study spectra of trained weight matrices, while the present derivation suggests a dynamical route by which anisotropic, activation-conditioned transport may accumulate in the weights. In particular, matrices of the form
\begin{equation}
G_\ell^{\alpha\beta}
=
\left(W^{(\ell+1)}\right)^TD_{\ell+1}^{\alpha\beta}W^{(\ell+1)}
\end{equation}
are not arbitrary diagnostics: they are the first nontrivial pullback operators required by the conjugate-field dynamics itself. This suggests that spectral concentration in trained weights may be linked to repeated gradient transport through co-activation-selected subspaces. The present paper does not derive a heavy-tailed law; it identifies the finite-depth transport mechanism whose spectra can be measured.

\section{Discussion}
\label{sec:discussion}

The main result is that gradient descent in fixed-readout ReLU networks admits a structured collective-field representation. The residual dynamics has a simple kernel form at every depth, but the internal field dynamics reveals more information than the residual kernel alone. Activation variations are organized by a push-forward response hierarchy, while conjugate-field variations are organized by a pullback hierarchy. The first nontrivial weight-induced Gram metric appears exactly when three hidden layers are present.

This clarifies the relation between shallow and deep feature learning. With one hidden layer, activation closure is immediate. With two hidden layers, the conjugate field becomes dynamical but remains governed by a scalar identity operator. With three hidden layers, deeper co-activation geometry is pulled back into earlier feature space through \(G_1\). At higher depth, iterated pullback transports generalize this mechanism.

Compared with NTK-like descriptions \cite{jacot2018ntk,roberts2022principles}, the emphasis here is not on a limiting kernel regime but on an exact first-order algebraic decomposition of the finite-depth update. Compared with empirical weight-spectrum diagnostics \cite{martin2021heavytailed,martin_weightwatcher}, the emphasis is not on post hoc spectral measurement but on the transport operators that arise directly in the learning dynamics. The recent analysis of weight Gram matrices in feature linearization \cite{cha2026weightgram} is closely related in spirit; the present work arrives at weight-induced Gram metrics through the recursive conjugate-field dynamics.

The derivation has several limitations. It is fixed-chamber and first-order in \(\eta\); finite learning rates and mask changes introduce additional terms. The readout is fixed, and all layers are assumed to have a common width for notational simplicity. Finally, the present paper deliberately stops at the algebraic collective-field level. Its role is to provide the fields, kernels, Gram metrics and cut-wise transport identities from which a later geometric formulation can be built.

\section{Conclusion}
\label{sec:conclusion}

We have refactored the learning dynamics of fixed-readout ReLU networks into a hierarchy of collective fields and transport operators. The explicit one-, two- and three-hidden-layer calculations show the successive emergence of activation closure, conjugate-field dynamics and the first pullback Gram metric. The arbitrary-depth formulas then organize the result into two dual recursions:
\begin{equation}
\cU_\ell
=
Q^{(\ell-1)}D_\ell+
\cF_{\ell-1}[\cU_{\ell-1}],
\qquad
\cM_\ell
=
S^{(\ell+1)}\Id+
\cP_\ell[\cM_{\ell+1}],
\end{equation}
and a residual kernel
\begin{equation}
K^{(L)}_{\alpha\beta}
=
\sum_{\ell=1}^{L}Q_{\alpha\beta}^{(\ell-1)}S_{\alpha\beta}^{(\ell)}.
\end{equation}
At every cut, the lower part of this kernel is reconstructed by the push-forward response \(\cU_m\), while the upper part is reconstructed by the pullback transport \(\cM_m\). This cut-wise duality is the central organizing principle of the paper.

The resulting picture is that depth does not merely add more nonlinear transformations; it builds a hierarchy of transport operators through which forward activation geometry and backward conjugate-field geometry are represented at different cuts. The first visible trace of this hierarchy is the weight-induced Gram metric \(G_\ell\), and its higher iterates encode progressively deeper transport. These are the collective-field results that the later geometric formulation will refine, without needing to introduce that later tensorial language here.

\bibliographystyle{plain}
\bibliography{references}

\end{document}